\documentclass[10pt]{article}
\usepackage[preprint]{tmlr}

\usepackage[utf8]{inputenc} % allow utf-8 input
\usepackage[T1]{fontenc}    % use 8-bit T1 fonts
\usepackage{hyperref}       % hyperlinks
\usepackage{url}            % simple URL typesetting
\usepackage{booktabs}       % professional-quality tables
\usepackage{amsfonts}       % blackboard math symbols
\usepackage{nicefrac}       % compact symbols for 1/2, etc.
\usepackage{microtype}      % microtypography
\usepackage{xcolor}         % colors
\usepackage{amsmath}
\usepackage{xcolor}
\usepackage{amsmath}
\usepackage{graphicx}
\usepackage{wrapfig}
\usepackage{bm}
\usepackage{pdflscape}
\usepackage{todonotes}
\usepackage{subcaption}
\usepackage[title]{appendix}
\usepackage{algorithm2e}

\title{Synthetic location trajectory generation using categorical diffusion models}

\author{%
  Simon Dirmeier\thanks{Correspondence to: \texttt{simon.dirmeier@sdsc.ethz.ch}}\\
  \addr  Swiss Data Science Center\\
  ETH Zurich, Switzerland\\
  \AND
  Ye Hong\\
  \addr ETH Zurich, Switzerland\\ 
  \AND
  Fernando Perez-Cruz\\
  \addr Swiss Data Science Center\\
  ETH Zurich, Switzerland\\
}

\def\month{MM}  % Insert correct month for camera-ready version
\def\year{YYYY} % Insert correct year for camera-ready version
\def\openreview{\url{https://openreview.net/forum?id=XXXX}} % Insert correct link to OpenReview for camera-ready version

\begin{document}

\maketitle

\begin{abstract}
Diffusion probabilistic models (DPMs) have rapidly evolved to be one of the predominant generative models for the simulation of synthetic data, for instance, for computer vision, audio, natural language processing, or biomolecule generation. Here, we propose using DPMs for the generation of synthetic individual location trajectories (ILTs) which are sequences of variables representing physical locations visited by individuals. ILTs are of major importance in mobility research to understand the mobility behavior of populations and to ultimately inform political decision-making. We represent ILTs as multi-dimensional categorical random variables and propose to model their joint distribution using a continuous DPM by first applying the diffusion process in a continuous unconstrained space and then mapping the continuous variables into a discrete space. We demonstrate that our model can synthesize realistic ILPs by comparing conditionally and unconditionally generated sequences to real-world ILPs from a GNSS tracking data set which suggests the potential use of our model for synthetic data generation, for example, for benchmarking models used in mobility research.
\end{abstract}

\section{Introduction}
The generation of realistic synthetic data, i.e., data with similar summary statistics as observational data, is of increasing importance in research fields where data privacy and preservation of confidentially are necessary. On the one hand, methodological, and consequently scientific, progress in a research field can get delayed or aggravated when due to data privacy no public benchmark data is available. For example, when a novel method is trained and evaluated on a restricted data set that cannot be shared, an objective comparison of methods is not possible, and research findings might not be representative of the performance of a method in general. On the other hand, an increased focus on experimental reproducibility requires being able to freely share data and methods among research facilities. These issues are particularly problematic in mobility research where data sets are, for instance, derived from individual GNSS tracking data or user surveys which can uniquely identify individuals and consequentially their daily behavior, such as their travelling patterns or shopping behaviors.

Recently, diffusion probabilistic models (DPMs; \cite{sohldickstein15deep,song2019generative,ho2020diffusion}) have emerged as novel paradigm in generative modelling outperforming state-of-the-art methods like generative adversarial networks or autoregressive models in the generation of images or sequential data, like audio or text \citep{dhariwal2021diffusion,kong2021diffwave,ho2022cascaded,saharia2022palette,gong2023diffuseq}. Apart from the fact that DPMs achieve good sample quality and coverage of the data distribution \citep{dhariwal2021diffusion}, they are exceptionally easy and robust to train using simple mean squared error objectives. DPMs are typically motivated by employing Gaussian transitions kernels that corrupt an initial data sample either in a sequence of discrete Markovian transitions or continuously in an SDE framework \citep{song2021scorebased,karras2022elucidating}, and then learning a reverse kernel to denoise this process. Modelling discrete data using DPMs has so far been conducted by applying diffusions on the probability simplex \citep{hoogeboom2021argmax,richemond2022categorical}, by embedding the data in an unconstrained space and then applying the diffusion process on the transformed variables \citep{li2022diffusion,strudel2022self,dieleman2022continuous}, autoregressively \citep{hoogeboom2022autoregressive},
via concrete score matching \citep{meng2022score}, or in the framework of stochastic differential equations \citep{richemond2022categorical,campbell2022continuous,sun2023scorebased}. In addition, several approaches have been introduced to, e.g., increase the sample quality \citep{nichol21improved,ho2022classifierfree,chen2023analog} or to reduce sampling times \citep{song2021denoising,cheng2022dpm}.

Here, we propose a method for synthetic generation of individual location trajectories (ILTs), i.e., sequences of visited locations of individuals among a set of fixed locations, using DPMs. Our method gracefully unifies recent developments in discrete diffusion and score-based modelling in a common framework and is capable of producing samples that to a high degree resemble real-world sequences. We evaluate our model on a real-world data set consisting of pre-processed GNSS tracking measurements and show that the model can a) produce realistic samples when conditioned on sub-sequences of observational ILTs and b) produce unconditionally generated sequences that are similar to real-world data in several key statistics employed as benchmark metrics in mobility research. 

The outline of the manuscript is as follows: Section~\ref{sec:methods} describes relevant background and methodology, 
Section~\ref{sec:results} introduces a model for synthetic ILT generation and presents experimental results, Section~\ref{sec:conclusion} summarizes our findings.

\section{Methods}
\label{sec:methods}
\subsection{Diffusion models}

Discrete-time diffusion probabilistic models (DPMs; \cite{sohldickstein15deep,ho2020diffusion,song2019generative}) are latent variable models defined by a \textit{reverse process} and a complementary \textit{forward process}. The forward process $q(\bm{y}_{1:T} | \bm{y}_0) = \prod_{t=1}^T q(\bm{y}_t | \bm{y}_{t - 1})$ corrupts an initial data point $\bm{y}_0 \sim q(\bm{y}_0)$ into a sequence of latent variables $\bm{y}_t$ using the transition kernels
\begin{equation*}
    q(\bm{y}_t | \bm{y}_{t - 1} ) = \mathcal{N} \left(\sqrt{1 - \beta_t}  \bm{y}_{t - 1}, \beta_t \bm{I} \right)
\end{equation*} 
The kernels are typically chosen to be Gaussians \citep{sohldickstein15deep,ho2020diffusion} but can depending on the data, for instance, also be uniform or discrete \citep{hoogeboom2021argmax,austin2021structured,johnson2021beyond}. The forward process is parameterized by a noise schedule $\beta_t$ that can be manually selected \citep{ho2020diffusion,nichol21improved,li2022diffusion} or learned during training \citep{kingma2021variational}. The reverse process $p_{\phi}(\bm{y}_{0:T}) =p(\bm{y}_{T})  \sum_{t=1}^T p_{\phi}(\bm{y}_{t - 1} | \bm{y}_{t})$ denoises the latent variables starting from $\bm{y}_T \sim p(\bm{y}_T)$ towards the data distribution $q(\bm{y}_0)$ where $p(\bm{y}_T)$ is a simple distribution that can be sampled from easily. Since the reverse transitions are typically intractable they are learned using a \textit{score model} $s_\phi(\bm{z}_t, t)$ with neural network weights $\phi$. Here, we parameterize the reverse transitions as
\begin{equation*}
p_{\phi}(\bm{y}_{t - 1} | \bm{y}_{t}) = \mathcal{N}\left(\mu_\phi \left( \bm{y}_t, t\right), \beta_t \bm{I}\right)
\end{equation*} 
where we discuss the parameterization of the reverse process mean $\mu_\phi(\bm{y}_t, t)$ below (Section~\ref{sec:parameterization}). The parameters of the score model can be optimized by score-matching \citep{hyvarinen05ascore,song2019generative} or by maximizing a lower bound to the marginal likelihood (ELBO; \cite{sohldickstein15deep,ho2020diffusion}):
\begin{equation}
\mathbb{E}_{q(\bm{y}_{1:T} | \bm{y}_0) } \biggl[ 
    \log p_\phi \left(\bm{y}_0 |\bm{y}_1 \right) 
    - \sum_{t=2}^T \mathbb{KL}\Bigl[ q(\bm{y}_{t - 1} |\bm{y}_{t}, \bm{y}_0), p_\phi(\bm{y}_{t - 1} |\bm{y}_t)  \Bigr]
    - \mathbb{KL}\Bigl[ q(\bm{y}_T |\bm{y}_0), p(\bm{y}_T)  \Bigr]
    \biggr]
\label{eqn:continuous-elbo}
\end{equation} 
In Equation~\eqref{eqn:continuous-elbo}, both the conditionals $q(\bm{y}_t | \bm{y}_0)$ and the forward process posteriors $q(\bm{y}_{t - 1} |\bm{y}_{t}, \bm{y}_0)$ be computed analytically via
\begin{equation*}
q(\bm{y}_t | \bm{y}_0) = \mathcal{N}\left(\sqrt{\bar{\alpha}_t}\bm{y}_{0},( 1 - \bar{\alpha}_t)\bm{I}\right) \quad \text{and} \quad q(\bm{y}_{t - 1} | \bm{y}_{t}, \bm{y}_0)  = \mathcal{N}\left(  \tilde{\mu}(\bm{y}_t, \bm{y}_0), \tilde{\beta}_t \bm{I} \right) 
\end{equation*} 
where $\alpha_t = 1 - \beta_t$, $\bar{\alpha}_t = \prod_{s=1}^t \alpha_s$, $\tilde{\beta}_t = \frac{1 - \bar{\alpha}_{t - 1}}{1 - \bar{\alpha}_t} \beta_t$ and
\begin{equation}
\tilde{\mu}(\bm{y}_t, \bm{y}_0)  = \frac{\sqrt{\bar{\alpha}_{t - 1}} \beta_t }{1 - \bar{\alpha}_t } \bm{y}_0 + \frac{\sqrt{\alpha_t} (1 - \bar{\alpha}_{t - 1})}{1 - \bar{\alpha}_t} \bm{y}_t
\label{eqn:params-forward}
\end{equation} 

\subsection{Categorical diffusions in continuous space}

The forward process of vanilla DPMs \citep{sohldickstein15deep, ho2020diffusion} assumes Gaussian forward kernels $q(\bm{y}_t | \bm{y}_{t - 1})$. In the case of ILTs, a sample $\bm{y}_0 = [{y}_{00}, {y}_{01}, \dots, {y}_{0N}]^T$ is a sequence of $N$ discrete variables of $D$ categories that represent trajectories of visited locations of an individual, i.e., analogously to the autoregressive nature of textual sequences. To make DPMs amendable to categorical data, we adopt the approach by \citet{li2022diffusion} and apply the diffusion process in a continuous space. Specifically, we first map the data $\bm{y}_0$ into an unconstrained latent space via an embedding function $\texttt{EMB}: \mathcal{Y} \rightarrow \mathbb{R}^P$ where $P$ is the dimensionality of the embedding. We then model $q(\bm{z}_0 | \bm{y}_0) = \mathcal{N} \left( \texttt{EMB}(\bm{y}_0), \sigma^2_0 \bm{I} \right)$ where in our experiments we set $\sigma^2_0 = \beta_1$. Finally, we apply the diffusion process to the noisy embedding $\bm{z}_0$. Complementarily, we define the reverse transition $p_\phi(\bm{y}_0 | \bm{z}_0)$ which we parameterize as $p_\phi(\bm{y}_0 | \bm{z}_0) = \prod_{n=1}^N p_\phi(y_{0n} | \bm{z}_{0n})$ where $p_\phi(y_{0n} | \bm{z}_{0n})$ is a categorical distribution (see Appendix~\ref{appendix:mathematical-details}).

Accounting for the embedding into a continuous space and applying the diffusion process there, leads to the following objective:
\begin{equation}
\begin{split}
\mathbb{E}_{q(\bm{z}_{0:T} | \bm{y}_0)} \biggl[& 
    \log p_\phi (\bm{y}_0 | \bm{z}_0)  - \log q(\bm{z}_0 | \bm{y}_0)\\
    &+ \underbrace{\log p_\phi \left(\bm{z}_0 |\bm{z}_1 \right)  - \sum_{t=2}^T \mathbb{KL}\Bigl[ q(\bm{z}_{t - 1} |\bm{z}_{t}, \bm{z}_0), p_\phi(\bm{z}_{t - 1} |\bm{z}_t)  \Bigr] - \mathbb{KL}\Bigl[ q(\bm{z}_T |\bm{z}_0), p(\bm{z}_T)  \Bigr]}_{\text{Objective within expectation of Equation~\ref{eqn:continuous-elbo}}}
    \biggr]
\label{eqn:full-continuous-elbo}
\end{split}
\end{equation} 
Equation~\eqref{eqn:full-continuous-elbo} replaces the diffusion on $\bm{y}_0$ with a diffusion on the latent variable $\bm{z}_0$. It can be further simplified and numerically stabilized by replacing the KL-terms with mean squared error terms
\begin{align*}
\mathbb{E}_{q(\bm{z}_{0:T} | \bm{y}_0)} \biggl[
    \log p_\phi (\bm{y}_0 | \bm{z}_0)  
     - || \texttt{EMB}(\bm{y}_0) - \mu_\phi(\bm{z}_1,  1) ||^2  
    - \sum_{t=2}^T ||   \tilde{\mu}(\bm{z}_t, \bm{z}_0) - \mu_\phi(\bm{z}_t, t) ||^2 
    - ||  \sqrt{\bar{\alpha}_T} \bm{z}_0 ||^2 
    \biggr]
\end{align*} 
The objective can be easily optimized using stochastic gradient ascent \citep{bottou2010large,hoffman2013stochastic,rezende14stochastic} (for more details we refer to Appendix~\ref{appendix:mathematical-details}). 

\subsection{Score model parameterization}
\label{sec:parameterization}

As described previously \citep{ho2020diffusion,song2019generative,kingma2021variational,li2022diffusion}, the score model $s_\phi(\bm{z}_t, t)$ can be used to either directly estimate the reverse process mean $\hat{\mu}_\phi(\bm{z}_t, t) \leftarrow s_\phi(\bm{z}_t, t)$, or to compute an estimate of the noise $\hat{\boldsymbol \epsilon}_{t} \leftarrow s_\phi(\bm{z}_t, t)$ that is used to corrupt $\bm{z}_{0}$ into $\bm{z}_t$, or an  estimate $\hat{\bm{z}}_{0} \leftarrow s_\phi(\bm{z}_t, t)$ of $\bm{z}_{0}$ itself.

The seminal work by \citet{ho2020diffusion} shows that when the score model is used to predict the noise $\boldsymbol \epsilon_t$, then an estimate of the original noisy embedding can be computed via 
\begin{equation}
\hat{\bm{z}}^t_0 \leftarrow \left( \bm{z}_t - \sqrt{1 - \bar{\alpha}_t} \hat{\boldsymbol \epsilon}_t \right) / \sqrt{\bar{\alpha}_t}
\end{equation}
 The estimate $\hat{\bm{z}}^t_0$ is then used to compute the reverse process mean via the formula of the forward process posterior (Equation~\eqref{eqn:params-forward}):
\begin{equation*}
    \hat{\mu}_\phi(\bm{z}_t, t) \leftarrow \tilde{\mu}(\bm{z}_t, \hat{\bm{z}}^t_0)
\end{equation*}

Here, we evaluate both approaches, i.e., prediction of an estimate of the noise $\boldsymbol{\epsilon}_t$ and prediction of an estimate of the noisy embedding ${\bm{z}}_0$. Prediction of the embedding instead of the noise can yield significant performance improvements in categorical diffusion modelling \citep{li2022diffusion,strudel2022self,chen2023analog} (see also Appendix~\ref{appendix:mathematical-details} for details).

\subsection{Self-conditioning}
\label{sec:selfcond}

We utilize \textit{self-conditioning} for improved performance of discrete diffusion models \citep{chen2023analog}. In the vanilla DPM implementation for the computation of the estimate $\hat{\mu}_\phi(\bm{z}_{t}, t)$, first an estimate $\hat{\bm{z}}^t_0$ is computed, then the formula of the forward posterior mean $\tilde{\mu}(\bm{z}_t, \hat{\bm{z}}^t_0)$ is employed (c.f. Section~\ref{sec:parameterization}), and finally a sample $\bm{z}_{t-1} \sim p_\phi \left(\tilde{\mu}(\bm{z}_t, \hat{\bm{z}}^t_0), \beta_t \bm{I} \right)$ is drawn. In the next reverse diffusion step, the previous computation $\hat{\bm{z}}^t_0$ is discarded. \citet{chen2023analog} propose to condition the score model $s_\phi(\bm{z}_t, t)$ additionally on its previous prediction as well, such that the new score model is parameterized as $s_\phi(\bm{z}_{t}, \hat{\bm{z}}^{t+1}_0, t)$. They show that this simple adaption can improve performance of diffusion models in discrete spaces significantly.

\subsection{Conditional synthesis}
\label{sec:conditional}

To generate location trajectories conditional on a sub-sequence, we make use of a simple masking procedure frequently used in NLP \citep{kenton2019bert,strudel2022self,dieleman2022continuous}.
 
During training, we mask $50\%$ of the rows of a noisy embedding at time $t$, $\bm{z}_t \in \mathbb{R}^{N \times P}$, by that effectively fixing some locations and only modelling the distribution over the others, and casting the conditional synthesis problem as an infilling problem \citep{donahue2020enabling}. Specifically, we construct the input to the score network by first stacking four sequences
\begin{itemize}
\item $\bm{z}_{t}$: the noisy embedding at time $t$, for which we set the rows $\bm{z}_{tn} = 0$ where $\bm{m}_{n}=1$,
\item $\hat{\bm{z}}^{t + 1}_{0}$: the estimate of the original noisy embedding at time $t + 1$,
\item $\bm{m}$: a binary mask of length $N$ of which $50\%$ of values are randomly set to $1$, indicating if a location is given ($\bm{m}_n=1$) or is to be generated ($\bm{m}_n=0$),
\item $\texttt{EMB}(\bm{y}_0)$: the original embedding matrix, where we analogously set the rows $\texttt{EMB}(\bm{y}_0)_{n} = \bm{0}$ where $\bm{m}_{n}=0$.
\end{itemize}
\citet{dieleman2022continuous} evaluate several masking strategies that could be adopted and demonstrate that random masking \citep{kenton2019bert} yields on average better objectives. We use a combination of \textit{prefix masking} where we mask the first $25\%$ of a latent embedding $\bm{z}_t$ and \textit{random masking} where we mask random elements of the rest of the embedding matrix (also $25\%$ of the total length $N$), such that in the end half of the location trajectory is masked and the other half is to be generated.

With this approach, we can synthesize sequences autoregressively, by providing a "prefix-seed" of fixed length and infilling the rest of the sequencing using a diffusion model.

\section{Synthetic location trajectory generation}
\label{sec:results}

We apply the categorical DPM (CDPM) to model the distribution of an ILT data set derived from the green class (GC) study. We briefly describe the data set, the score model used for the reverse diffusion process and several ablations we conducted to identify the best architecture, and then present experimental results where we compare statistics of synthetic ILTs to real-world ILTs.

\subsection{Data}

We use longitudinal GNSS tracking data derived from the GC study aiming at evaluating the impact of a mobility offer on the mobility behavior of individuals~\citep{martin2019green}. 
% stay points
Raw GNSS measurements underwent preprocessing to identify and delineate stationary areas of individuals, referred to as \textit{stay points}. 
% locations
Subsequently, these stay points are aggregated spatially to create \textit{locations}, denoted as $y_n$, to account for GNSS recording errors that may occur when individuals visit the same place. 
% user selection
For each participant, we segmented their location trajectory into subsequences of length $32$. For further details, we refer to Appendix~\ref{appendix:additional-details}.

\subsection{Model}
\label{sec:model}

We use a transformer architecture \citep{vaswani2017attention,kenton2019bert} as the core of our model $s_\phi(\cdot)$ in lieu of the conventional U-Net architectures \citep{ronneberger2015u} which are typically used for images. The transformer takes as input the a matrix consisting of a noisy embedding matrix $\bm{z}_t$, a prior-estimate $\hat{\bm{z}}_0$ found through self-conditioning, the original embedding $\texttt{EMB}(\bm{y}_0)$, and a mask $\bm{m}$ (c.f Section~\ref{sec:conditional}; see Appendix~\ref{appendix:additional-details} for details on architecture and training).

\begin{wrapfigure}{r}{0.5\textwidth}
\centering
\includegraphics[width=.5\textwidth]{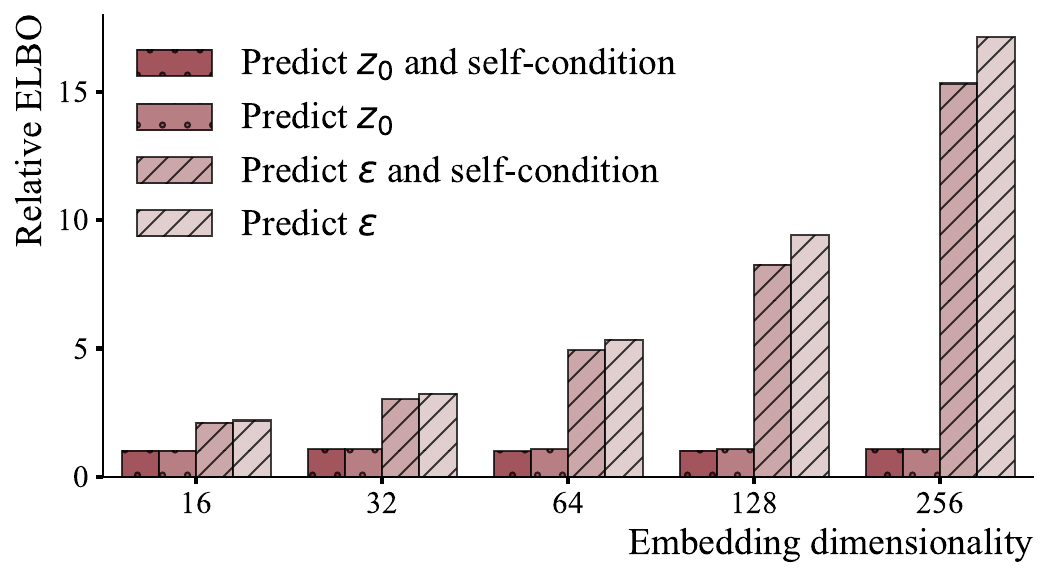}
\caption{Ablation study. We evaluate the influence of embedding dimensionality, score model parameterization and use of self-conditioning on the objective function (Equation~\ref{eqn:full-continuous-elbo}) and show the value of the objectives on a validation set relative to the best parameterization (i.e., best model has a relative ELBO of $1$, all others are higher). We find that an embedding dimensionality of $16$ with $\bm{z}_0$-parameterization and self-conditioning yields the best results on our data set, and that the embedding dimensionality has only minimal influence on performance with a $\bm{z}_0$-parameterization in comparison to the ${\boldsymbol \epsilon}_t$-parameterization (c.f. \citet{li2022diffusion}).}
\label{fig:architecture-ablation}
\end{wrapfigure} 

We conducted an ablation study to determine a) the optimal embedding dimensionality (i.e., the dimensionality produced by transforming $\texttt{EMB}(\bm{y}_0)$), b) whether to use self-conditioning or not, and c) whether to have the score model predict estimates $\hat{\bm{z}}_0$ or $\hat{\boldsymbol \epsilon_t}$, respectively. We find that the model that uses an embedding dimensionality of $16$, self-conditioning and parameterization to predict $\hat{\bm{z}}_0$ overall outperforms the other parameterizations (Figure ~\ref{fig:architecture-ablation}; more ablations on optimal time-step embedding dimensionality, noise schedule and classifier-free guidance \citep{ho2022classifierfree} can be found in Appendix~\ref{appendix:additional-results}).

\subsection{Baselines}

We compare the model against several mechanistic baselines from the mobility literature, that are usually employed for synthetic data generation: EPR, dEPR, dtEPR, IPT (we refer to \cite{barbosa2018human}, \cite{hong2023revealing} and references therein for descriptions of these models).

Note that all of the baselines above require access to observational data to simulate synthetic sequences while a data-driven approach like the CDPM only requires access once when training the model.

\subsection{Experimental results}

\begin{figure}
\centering
 \begin{subfigure}[b]{1\textwidth}
 \caption{Sequence entropies.}
 \includegraphics[width=1\textwidth]{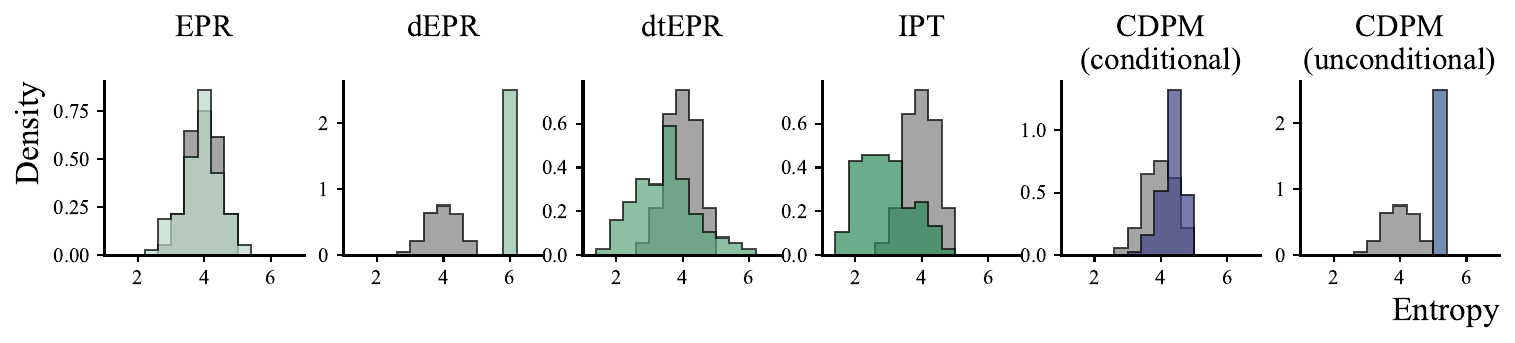}
 \end{subfigure}
 \begin{subfigure}[b]{1\textwidth}
 \caption{Number of visits per location.}
 \includegraphics[width=1\textwidth]{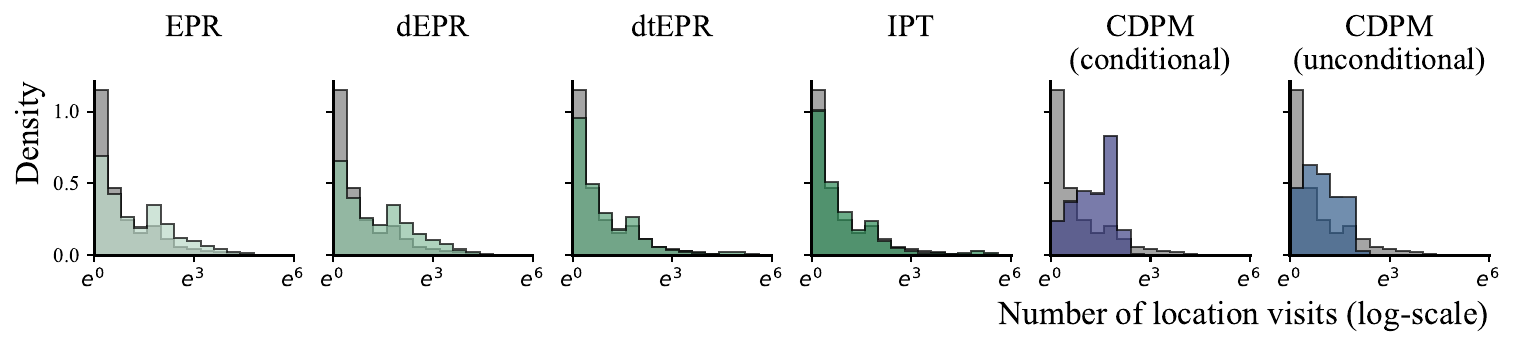}
 \end{subfigure}
 \begin{subfigure}[b]{1\textwidth}
 \caption{Distances between pairs of consecutive locations.}
 \includegraphics[width=1\textwidth]{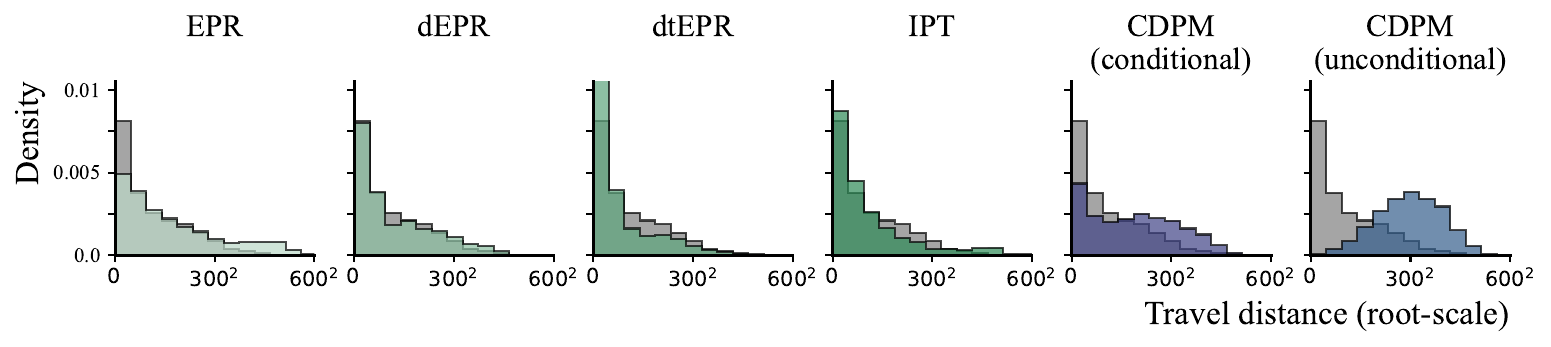}
 \end{subfigure}
\caption{Empirical comparison of statistics of observational and synthesized location sequences. We compute the entropy, the number of visits per location and travel distances over each location trajectory and visualize the histogram of these statistics (the greater the overlap the better). The first four columns in each figure show mechanistic models, the last two columns show CDPM simulations with conditional and unconditional synthesis, respectively.}
\label{fig:comparison-statistics}
\end{figure} 

Since in mobility research, data sets can often not be shared due to privacy protocols, we evaluate synthesis of conditionally and unconditionally generated sequences on several privacy-conserving metrics frequently used for synthetic ILT generation \citep{feng2020learning}. 

In the conditional case, we iteratively use a seed sequence of $50\%$ of the length of the modelled output which we take to be $N=32$ and infill the other $50\%$ and then use the infilled sequence in an autoregressive manner as seed for the next iteration. In the unconditional case, we do not provide any initial seed, but adopt the same autoregressive data generation. We use the following metrics for evaluation:

\begin{itemize}
    \item \textbf{Shannon entropy}: we evaluate the Shannon entropy on each generated and real-world sequence and compare their empirical distributions.
    \item \textbf{Number of visits per location}: for each generated and real-world sequence, we compute the number of visits per each unique location.
    \item \textbf{Travel distances}: for each generated sequence, we compute the distance travelled between each pair of consecutive locations. To do so, we map the synthesized locations back to their original GNSS tracking locations and then compute the distances between the GNSS locations. In the conditional scenario, since we start a trajectory with a seed sequence that represents real-world locations, the subsequent locations should correspond to the same locations as in the real-world data.
\end{itemize}

We find that in the conditional case (Figure~\ref{fig:comparison-statistics}), the CDPM can simulate sequences that adequately represent statistics of the observational data. While the computed entropies are sufficiently close to real-world data, our model slightly underestimates the mean and variance of the empirical entropies but overall seems to represent the original data closely. This can be explained by the fact that our model simulates sequences where the frequency of number of visits per location have a lower tail in comparison to the observational data. In comparison to the empirical travel distances, the distribution of synthetic distances is overestimated by our model. Since we do not directly model the distances between locations, but only learn these from the sequence categories, failing to model these adequately is expected. 

In the unconditional case (Figure~\ref{fig:comparison-statistics}), both entropies and number of location visits have similar statistics as in the conditional case. The estimated travel distances in this case, however, are distributed more uniformly which we attribute to the fact that the initial seed sequence does not correspond to a previously seen location trajectory, but instead consists of a set of random initial locations. 

In contrast, while the mechanistic models are able to simulate synthetic location trajectories w.r.t. number of visits per location and pair-wise distances, they fail to simulate sequences with similar entropies as the observational data.

\section{Conclusion and limitations}
\label{sec:conclusion}
We presented a model based on categorical DPMs for synthetic generation of individual location trajectories. Our method builds on prior work by \citet{li2022diffusion} who presented a language model for text generation that uses latent diffusions to model the embedding matrix of word embeddings. In comparison to their work, which requires using an additional neural network classifier, we use a simple masking mechanism and treat the conditional generation problem as an infilling problem. In addition, we extend the model with self-conditioning and classifier-free guidance.

We trained a model using location sequences of length $N=32$ with relatively low sample size in comparison to the input dimensionality. We found that in our experiments this necessitated to have a relatively high number of masked locations (here $50\%$, compare to the $15\%$ in \citet{kenton2019bert}) which arguably reduces variability in the simulated trajectories. We suspect this can easily be resolved by training the model on a larger data set and use this model for benchmarking novel methods in mobility research, or by reducing the mask to, e.g., $25\%$ during training and sampling.

Especially in research fields that make use of confidential data sets of individuals such as mobility research, medicine or psychology, methods that can generate realistic data are crucial to facilitate, for instance, objective validation of novel methodology and establishment of common benchmark data, and we hope our contribution to be valuable to the mobility research community.

\section*{Acknowledgments and Disclosure of Funding}

This work was supported by the Hasler Foundation under the project titled "Interpretable and Robust Machine Learning for Mobility Analysis" (grant number 21041).

\newpage
\bibliographystyle{plainnat}
\bibliography{references} 

\begin{thebibliography}{44}
\providecommand{\natexlab}[1]{#1}
\providecommand{\url}[1]{\texttt{#1}}
\expandafter\ifx\csname urlstyle\endcsname\relax
  \providecommand{\doi}[1]{doi: #1}\else
  \providecommand{\doi}{doi: \begingroup \urlstyle{rm}\Url}\fi

\bibitem[Austin et~al.(2021)Austin, Johnson, Ho, Tarlow, and van~den
  Berg]{austin2021structured}
Jacob Austin, Daniel~D. Johnson, Jonathan Ho, Daniel Tarlow, and Rianne van~den
  Berg.
\newblock Structured denoising diffusion models in discrete state-spaces.
\newblock In \emph{Advances in Neural Information Processing Systems}, 2021.

\bibitem[Babuschkin et~al.(2020)Babuschkin, Baumli, Bell, Bhupatiraju, Bruce,
  Buchlovsky, Budden, Cai, Clark, Danihelka, Dedieu, Fantacci,
  et~al.]{deepmind2020jax}
Igor Babuschkin, Kate Baumli, Alison Bell, Surya Bhupatiraju, Jake Bruce, Peter
  Buchlovsky, David Budden, Trevor Cai, Aidan Clark, Ivo Danihelka, Antoine
  Dedieu, Claudio Fantacci, et~al.
\newblock The {D}eep{M}ind {JAX} {E}cosystem, 2020.
\newblock URL \url{http://github.com/deepmind}.

\bibitem[Barbosa et~al.(2018)Barbosa, Barthelemy, Ghoshal, James, Lenormand,
  Louail, Menezes, Ramasco, Simini, and Tomasini]{barbosa2018human}
Hugo Barbosa, Marc Barthelemy, Gourab Ghoshal, Charlotte~R James, Maxime
  Lenormand, Thomas Louail, Ronaldo Menezes, Jos{\'e}~J Ramasco, Filippo
  Simini, and Marcello Tomasini.
\newblock Human mobility: Models and applications.
\newblock \emph{Physics Reports}, 734:\penalty0 1--74, 2018.

\bibitem[Bottou(2010)]{bottou2010large}
L{\'e}on Bottou.
\newblock Large-scale machine learning with stochastic gradient descent.
\newblock In \emph{Proceedings of COMPSTAT'2010}, pages 177--186, 2010.

\bibitem[Campbell et~al.(2022)Campbell, Benton, De~Bortoli, Rainforth,
  Deligiannidis, and Doucet]{campbell2022continuous}
Andrew Campbell, Joe Benton, Valentin De~Bortoli, Thomas Rainforth, George
  Deligiannidis, and Arnaud Doucet.
\newblock A continuous time framework for discrete denoising models.
\newblock \emph{Advances in Neural Information Processing Systems}, 2022.

\bibitem[Chen et~al.(2023)Chen, Zhang, and Hinton]{chen2023analog}
Ting Chen, Ruixiang Zhang, and Geoffrey Hinton.
\newblock Analog bits: Generating discrete data using diffusion models with
  self-conditioning.
\newblock In \emph{The Eleventh International Conference on Learning
  Representations}, 2023.

\bibitem[Devlin et~al.(2019)Devlin, Chang, Lee, and Toutanova]{kenton2019bert}
Jacob Devlin, Ming{-}Wei Chang, Kenton Lee, and Kristina Toutanova.
\newblock {BERT:} pre-training of deep bidirectional transformers for language
  understanding.
\newblock In \emph{Proceedings of the 2019 Conference of the North American
  Chapter of the Association for Computational Linguistics: Human Language
  Technologies}, pages 4171--4186, 2019.

\bibitem[Dhariwal and Nichol(2021)]{dhariwal2021diffusion}
Prafulla Dhariwal and Alexander~Quinn Nichol.
\newblock Diffusion models beat {GAN}s on image synthesis.
\newblock In \emph{Advances in Neural Information Processing Systems}, 2021.

\bibitem[Dieleman et~al.(2022)Dieleman, Sartran, Roshannai, Savinov, Ganin,
  Richemond, Doucet, Strudel, Dyer, Durkan, et~al.]{dieleman2022continuous}
Sander Dieleman, Laurent Sartran, Arman Roshannai, Nikolay Savinov, Yaroslav
  Ganin, Pierre~H Richemond, Arnaud Doucet, Robin Strudel, Chris Dyer, Conor
  Durkan, et~al.
\newblock Continuous diffusion for categorical data.
\newblock \emph{arXiv preprint arXiv:2211.15089}, 2022.

\bibitem[Donahue et~al.(2020)Donahue, Lee, and Liang]{donahue2020enabling}
Chris Donahue, Mina Lee, and Percy Liang.
\newblock Enabling language models to fill in the blanks.
\newblock In \emph{Proceedings of the 58th Annual Meeting of the Association
  for Computational Linguistics}, 2020.

\bibitem[Feng et~al.(2020)Feng, Yang, Xu, Yu, Wang, and Li]{feng2020learning}
Jie Feng, Zeyu Yang, Fengli Xu, Haisu Yu, Mudan Wang, and Yong Li.
\newblock Learning to simulate human mobility.
\newblock In \emph{Proceedings of the 26th ACM SIGKDD international conference
  on knowledge discovery \& data mining}, pages 3426--3433, 2020.

\bibitem[Gong et~al.(2023)Gong, Li, Feng, Wu, and Kong]{gong2023diffuseq}
Shansan Gong, Mukai Li, Jiangtao Feng, Zhiyong Wu, and Lingpeng Kong.
\newblock {DiffuSeq}: Sequence to sequence text generation with diffusion
  models.
\newblock In \emph{The Eleventh International Conference on Learning
  Representations}, 2023.

\bibitem[Hendrycks and Gimpel(2016)]{hendrycks2016gaussian}
Dan Hendrycks and Kevin Gimpel.
\newblock Gaussian error linear units ({GELU}s).
\newblock \emph{arXiv preprint arXiv:1606.08415}, 2016.

\bibitem[Hennigan et~al.(2020)Hennigan, Cai, Norman, Martens, and
  Babuschkin]{haiku2020github}
Tom Hennigan, Trevor Cai, Tamara Norman, Lena Martens, and Igor Babuschkin.
\newblock {H}aiku: {S}onnet for {JAX}, 2020.
\newblock URL \url{http://github.com/deepmind/dm-haiku}.

\bibitem[Ho and Salimans(2022)]{ho2022classifierfree}
Jonathan Ho and Tim Salimans.
\newblock Classifier-free diffusion guidance.
\newblock \emph{arXiv preprint arXiv:2207.12598}, 2022.

\bibitem[Ho et~al.(2020)Ho, Jain, and Abbeel]{ho2020diffusion}
Jonathan Ho, Ajay Jain, and Pieter Abbeel.
\newblock Denoising diffusion probabilistic models.
\newblock In \emph{Advances in Neural Information Processing Systems}, 2020.

\bibitem[Ho et~al.(2022)Ho, Saharia, Chan, Fleet, Norouzi, and
  Salimans]{ho2022cascaded}
Jonathan Ho, Chitwan Saharia, William Chan, David~J Fleet, Mohammad Norouzi,
  and Tim Salimans.
\newblock Cascaded diffusion models for high fidelity image generation.
\newblock \emph{The Journal of Machine Learning Research}, 23\penalty0
  (1):\penalty0 2249--2281, 2022.

\bibitem[Hoffman et~al.(2013)Hoffman, Blei, Wang, and
  Paisley]{hoffman2013stochastic}
Matthew~D Hoffman, David~M Blei, Chong Wang, and John Paisley.
\newblock Stochastic variational inference.
\newblock \emph{Journal of Machine Learning Research}, 2013.

\bibitem[Hong et~al.(2023)Hong, Xin, Dirmeier, Perez-Cruz, and
  Raubal]{hong2023revealing}
Ye~Hong, Yanan Xin, Simon Dirmeier, Fernando Perez-Cruz, and Martin Raubal.
\newblock Revealing behavioral impact on mobility prediction networks through
  causal interventions.
\newblock \emph{arXiv preprint arXiv:2311.11749}, 2023.

\bibitem[Hoogeboom et~al.(2021)Hoogeboom, Nielsen, Jaini, Forr\'{e}, and
  Welling]{hoogeboom2021argmax}
Emiel Hoogeboom, Didrik Nielsen, Priyank Jaini, Patrick Forr\'{e}, and Max
  Welling.
\newblock Argmax flows and multinomial diffusion: Learning categorical
  distributions.
\newblock In \emph{Advances in Neural Information Processing Systems}, 2021.

\bibitem[Hoogeboom et~al.(2022)Hoogeboom, Gritsenko, Bastings, Poole, van~den
  Berg, and Salimans]{hoogeboom2022autoregressive}
Emiel Hoogeboom, Alexey~A. Gritsenko, Jasmijn Bastings, Ben Poole, Rianne
  van~den Berg, and Tim Salimans.
\newblock Autoregressive diffusion models.
\newblock In \emph{International Conference on Learning Representations}, 2022.

\bibitem[Hyv{{\"a}}rinen(2005)]{hyvarinen05ascore}
Aapo Hyv{{\"a}}rinen.
\newblock Estimation of non-normalized statistical models by score matching.
\newblock \emph{Journal of Machine Learning Research}, 6\penalty0
  (24):\penalty0 695--709, 2005.

\bibitem[Johnson et~al.(2021)Johnson, Austin, van~den Berg, and
  Tarlow]{johnson2021beyond}
Daniel~D. Johnson, Jacob Austin, Rianne van~den Berg, and Daniel Tarlow.
\newblock Beyond in-place corruption: Insertion and deletion in denoising
  probabilistic models.
\newblock In \emph{ICML Workshop on Invertible Neural Networks, Normalizing
  Flows, and Explicit Likelihood Models}, 2021.

\bibitem[Karras et~al.(2022)Karras, Aittala, Aila, and
  Laine]{karras2022elucidating}
Tero Karras, Miika Aittala, Timo Aila, and Samuli Laine.
\newblock Elucidating the design space of diffusion-based generative models.
\newblock In \emph{Advances in Neural Information Processing Systems}, 2022.

\bibitem[Kingma et~al.(2021)Kingma, Salimans, Poole, and
  Ho]{kingma2021variational}
Diederik Kingma, Tim Salimans, Ben Poole, and Jonathan Ho.
\newblock Variational diffusion models.
\newblock In \emph{Advances in neural information processing systems}, 2021.

\bibitem[Kong et~al.(2021)Kong, Ping, Huang, Zhao, and
  Catanzaro]{kong2021diffwave}
Zhifeng Kong, Wei Ping, Jiaji Huang, Kexin Zhao, and Bryan Catanzaro.
\newblock {DiffWave}: A versatile diffusion model for audio synthesis.
\newblock In \emph{International Conference on Learning Representations}, 2021.

\bibitem[Li et~al.(2022)Li, Thickstun, Gulrajani, Liang, and
  Hashimoto]{li2022diffusion}
Xiang~Lisa Li, John Thickstun, Ishaan Gulrajani, Percy Liang, and Tatsunori
  Hashimoto.
\newblock Diffusion-{LM} improves controllable text generation.
\newblock In \emph{Advances in Neural Information Processing Systems}, 2022.

\bibitem[Loshchilov and Hutter(2019)]{loshchilov2018decoupled}
Ilya Loshchilov and Frank Hutter.
\newblock Decoupled weight decay regularization.
\newblock In \emph{International Conference on Learning Representations}, 2019.

\bibitem[Lu et~al.(2022)Lu, Zhou, Bao, Chen, LI, and Zhu]{cheng2022dpm}
Cheng Lu, Yuhao Zhou, Fan Bao, Jianfei Chen, Chongxuan LI, and Jun Zhu.
\newblock Dpm-solver: A fast ode solver for diffusion probabilistic model
  sampling in around 10 steps.
\newblock In \emph{Advances in Neural Information Processing Systems}, 2022.

\bibitem[Martin et~al.(2019)Martin, Becker, Bucher, Jonietz, Raubal, and
  Axhausen]{martin2019green}
Henry Martin, Henrik Becker, Dominik Bucher, David Jonietz, Martin Raubal, and
  Kay~W. Axhausen.
\newblock Begleitstudie {SBB Green Class} - {A}bschlussbericht.
\newblock 2019.

\bibitem[Meng et~al.(2022)Meng, Choi, Song, and Ermon]{meng2022score}
Chenlin Meng, Kristy Choi, Jiaming Song, and Stefano Ermon.
\newblock Concrete score matching: Generalized score matching for discrete
  data.
\newblock In \emph{Advances in Neural Information Processing Systems}, 2022.

\bibitem[Nichol and Dhariwal(2021)]{nichol21improved}
Alexander~Quinn Nichol and Prafulla Dhariwal.
\newblock Improved denoising diffusion probabilistic models.
\newblock In \emph{Proceedings of the 38th International Conference on Machine
  Learning}, 2021.

\bibitem[Rezende et~al.(2014)Rezende, Mohamed, and
  Wierstra]{rezende14stochastic}
Danilo~Jimenez Rezende, Shakir Mohamed, and Daan Wierstra.
\newblock Stochastic backpropagation and approximate inference in deep
  generative models.
\newblock In \emph{Proceedings of the 31st International Conference on Machine
  Learning}, 2014.

\bibitem[Richemond et~al.(2022)Richemond, Dieleman, and
  Doucet]{richemond2022categorical}
Pierre~H Richemond, Sander Dieleman, and Arnaud Doucet.
\newblock Categorical {SDE}s with simplex diffusion.
\newblock \emph{arXiv preprint arXiv:2210.14784}, 2022.

\bibitem[Ronneberger et~al.(2015)Ronneberger, Fischer, and
  Brox]{ronneberger2015u}
Olaf Ronneberger, Philipp Fischer, and Thomas Brox.
\newblock U-net: Convolutional networks for biomedical image segmentation.
\newblock In \emph{Medical Image Computing and Computer-Assisted
  Intervention--MICCAI 2015: 18th International Conference, Munich, Germany,
  October 5-9, 2015, Proceedings, Part III 18}, pages 234--241. Springer, 2015.

\bibitem[Saharia et~al.(2022)Saharia, Chan, Chang, Lee, Ho, Salimans, Fleet,
  and Norouzi]{saharia2022palette}
Chitwan Saharia, William Chan, Huiwen Chang, Chris Lee, Jonathan Ho, Tim
  Salimans, David Fleet, and Mohammad Norouzi.
\newblock Palette: Image-to-image diffusion models.
\newblock In \emph{ACM SIGGRAPH 2022 Conference Proceedings}, pages 1--10,
  2022.

\bibitem[Sohl-Dickstein et~al.(2015)Sohl-Dickstein, Weiss, Maheswaranathan, and
  Ganguli]{sohldickstein15deep}
Jascha Sohl-Dickstein, Eric Weiss, Niru Maheswaranathan, and Surya Ganguli.
\newblock Deep unsupervised learning using nonequilibrium thermodynamics.
\newblock In \emph{Proceedings of the 32nd International Conference on Machine
  Learning}, 2015.

\bibitem[Song et~al.(2021{\natexlab{a}})Song, Meng, and
  Ermon]{song2021denoising}
Jiaming Song, Chenlin Meng, and Stefano Ermon.
\newblock Denoising diffusion implicit models.
\newblock In \emph{International Conference on Learning Representations},
  2021{\natexlab{a}}.

\bibitem[Song and Ermon(2019)]{song2019generative}
Yang Song and Stefano Ermon.
\newblock Generative modeling by estimating gradients of the data distribution.
\newblock In \emph{Advances in Neural Information Processing Systems}, 2019.

\bibitem[Song et~al.(2021{\natexlab{b}})Song, Sohl-Dickstein, Kingma, Kumar,
  Ermon, and Poole]{song2021scorebased}
Yang Song, Jascha Sohl-Dickstein, Diederik~P Kingma, Abhishek Kumar, Stefano
  Ermon, and Ben Poole.
\newblock Score-based generative modeling through stochastic differential
  equations.
\newblock In \emph{International Conference on Learning Representations},
  2021{\natexlab{b}}.

\bibitem[Strudel et~al.(2022)Strudel, Tallec, Altch{\'e}, Du, Ganin, Mensch,
  Grathwohl, Savinov, Dieleman, Sifre, et~al.]{strudel2022self}
Robin Strudel, Corentin Tallec, Florent Altch{\'e}, Yilun Du, Yaroslav Ganin,
  Arthur Mensch, Will Grathwohl, Nikolay Savinov, Sander Dieleman, Laurent
  Sifre, et~al.
\newblock Self-conditioned embedding diffusion for text generation.
\newblock \emph{arXiv preprint arXiv:2211.04236}, 2022.

\bibitem[Sun et~al.(2023)Sun, Yu, Dai, Schuurmans, and Dai]{sun2023scorebased}
Haoran Sun, Lijun Yu, Bo~Dai, Dale Schuurmans, and Hanjun Dai.
\newblock Score-based continuous-time discrete diffusion models.
\newblock In \emph{The Eleventh International Conference on Learning
  Representations}, 2023.

\bibitem[Vaswani et~al.(2017)Vaswani, Shazeer, Parmar, Uszkoreit, Jones, Gomez,
  Kaiser, and Polosukhin]{vaswani2017attention}
Ashish Vaswani, Noam Shazeer, Niki Parmar, Jakob Uszkoreit, Llion Jones,
  Aidan~N Gomez, {\L}ukasz Kaiser, and Illia Polosukhin.
\newblock Attention is all you need.
\newblock \emph{Advances in Neural Information Processing Systems}, 2017.

\bibitem[Xiong et~al.(2020)Xiong, Yang, He, Zheng, Zheng, Xing, Zhang, Lan,
  Wang, and Liu]{xiong2020prelayer}
Ruibin Xiong, Yunchang Yang, Di~He, Kai Zheng, Shuxin Zheng, Chen Xing,
  Huishuai Zhang, Yanyan Lan, Liwei Wang, and Tieyan Liu.
\newblock On layer normalization in the transformer architecture.
\newblock In \emph{Proceedings of the 37th International Conference on Machine
  Learning}, 2020.

\end{thebibliography}

\newpage
\begin{appendices}
\section{Mathematical details}
\label{appendix:mathematical-details}

\subsection{Likelihood}

We parameterize the likelihood $p_\phi(\bm{y}_0|\bm{z}_0) = \prod_{n=1}^N p_\phi({y}_{0n}|\bm{z}_{0n}) $ as $p_\phi({y}_{0n}|\bm{z}_{0n}) = \text{Categorical}(\boldsymbol \alpha_n)$ which we parameterize using logits $\boldsymbol \alpha_n = \bm{z}_{0n} \bm{E}^T$ where $\bm{E} \in \mathbb{R}^{D \times P}$ is the weight matrix of the embedding function $\texttt{EMB}(\cdot)$. 

\subsection{Training objective}

Equation~\eqref{eqn:full-continuous-elbo} can be further simplified by replacing the likelihood term with a cross-entropy term $\ell_c(\bm{y}_0, \bm{z}_0)$

\begin{align*}
 \mathbb{E}_{q(\bm{z}_{0:T} | \bm{y}_0)} \biggl[
- \ell_c(\bm{y}_0, \bm{z}_0)
     - || \texttt{EMB}(\bm{y}_0) - \mu_\phi(\bm{z}_1, 1) ||^2  
    - \sum_{t=2}^T ||   \tilde{\mu}(\bm{z}_t, \bm{z}_0) - \mu_\phi(\bm{z}_t, t) ||^2 
    - ||  \sqrt{\bar{\alpha}_T} \bm{z}_0 ||^2 
    \biggr] 
\end{align*} 

such that entire object does not require evaluating probability densities. 

In practice, instead of evaluating the objective on the entire chain $\bm{z}_{0:T} \sim q(\bm{z}_{0:T} | \bm{y}_0)$, we sample only a single 
$\bm{z}_{t} \sim q(\bm{z}_{t} | \bm{z}_0)$ and use the objective

\begin{align*}
 \mathbb{E}_{q(\bm{z}_{0,1, t} | \bm{y}_0)} \biggl[
- \ell_c(\bm{y}_0, \bm{z}_0)
     - || \texttt{EMB}(\bm{y}_0) - \mu_\phi(\bm{z}_1, 1) ||^2  
    - ||   \tilde{\mu}(\bm{z}_t, \bm{z}_0) - \mu_\phi(\bm{z}_t, t) ||^2 
    - ||  \sqrt{\bar{\alpha}_T} \bm{z}_0 ||^2 
    \biggr] 
\end{align*} 

which is significantly faster to evaluate.

\subsection{Parameterization}

In the case where the score model is parameterized to predict an estimate of the original noise embedding $\bm{z}_0$, we use the following objective

\begin{align*}
 \mathbb{E}_{q(\bm{z}_{0:T} | \bm{y}_0)} \biggl[
- \ell_c(\bm{y}_0, \bm{z}_0)
     - || \texttt{EMB}(\bm{y}_0) - s_\phi(\bm{z}_1, 1, \cdot) ||^2  
    - \sum_{t=2}^T ||  \bm{z}_0 - s_\phi(\bm{z}_t, t, \cdot) ||^2 
    - ||  \sqrt{\bar{\alpha}_T} \bm{z}_0 ||^2 
    \biggr] 
\end{align*} 

where we used the shorthand $s_\phi(\bm{z}_t, t, \cdot) = s_\phi(\bm{z}_t, t, \hat{\bm{z}}^t_0, \texttt{EMB}(\bm{y}_0), \bm{m})$.

Analogously, if the score model is parameterized to predict the noise $\bm{\epsilon}_t$ we optimize

\begin{align*}
 \mathbb{E}_{q(\bm{z}_{0:T} | \bm{y}_0)} \biggl[
- \ell_c(\bm{y}_0, \bm{z}_0)
     - || \texttt{EMB}(\bm{y}_0) - \hat{\bm{z}}^t_0 ||^2  
    - \sum_{t=2}^T ||  \boldsymbol \epsilon_t - s_\phi(\bm{z}_t, t, \cdot) ||^2 
    - ||  \sqrt{\bar{\alpha}_T} \bm{z}_0 ||^2 
    \biggr] 
\end{align*} 

where we compute $\hat{\bm{z}}^t_0$ via the reparameterization 

\begin{equation*}
\hat{\bm{z}}^t_0 \leftarrow \left( \bm{z}_t - \sqrt{1 - \bar{\alpha}_t} s_\phi(\bm{z}_t, t, \cdot) \right) / \sqrt{\bar{\alpha}_t} 
\end{equation*}

\section{Experimental details}
\label{appendix:additional-details}

\subsection{Model architecture}

\begin{figure}[h!]
\centering
\includegraphics[width=0.5\textwidth]{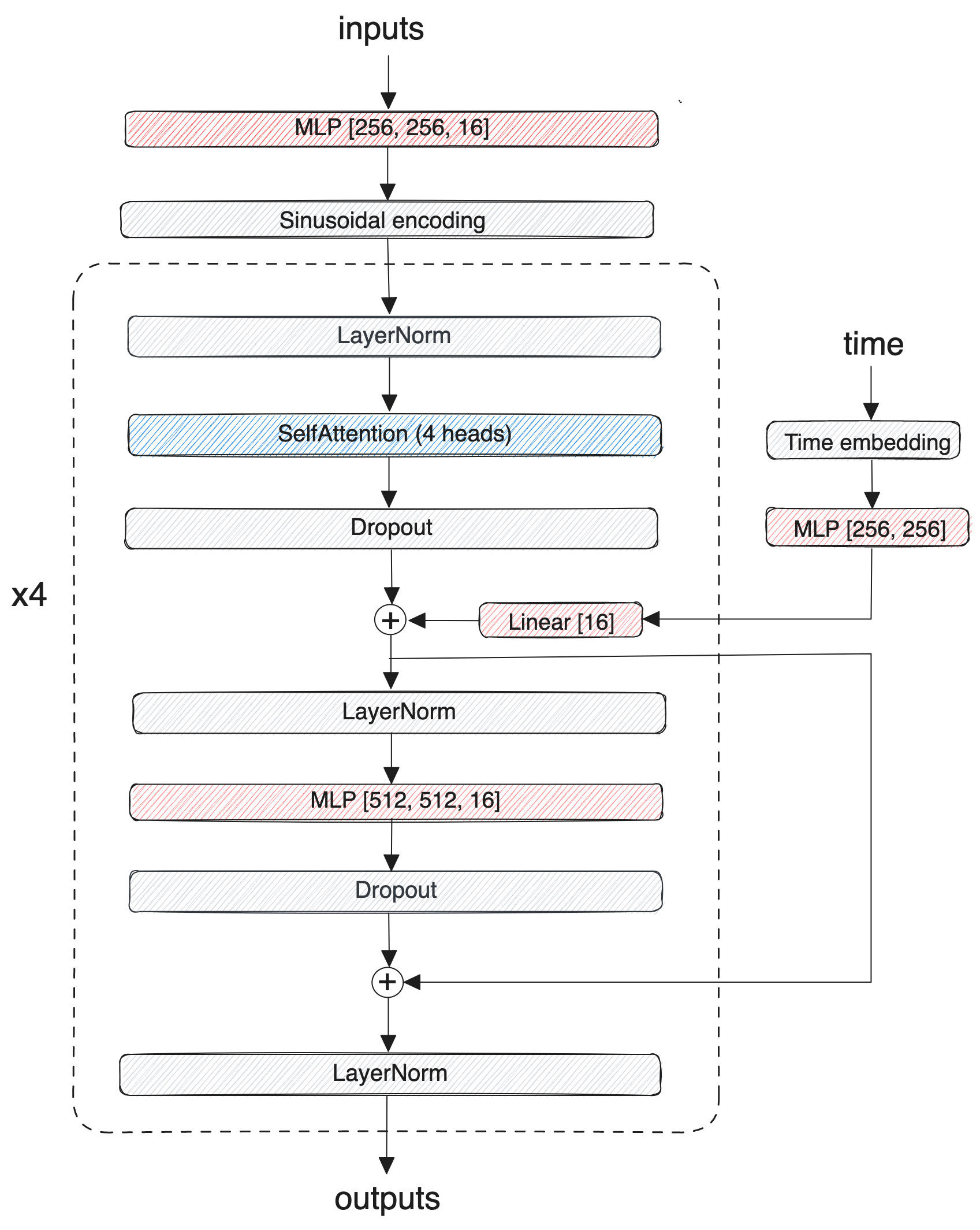}
\caption{Model architecture.}
\label{fig:architecture}
\end{figure}

We use a modified pre-LayerNorm transformer architecture \citep{xiong2020prelayer} as a score model $s_\phi(\bm{z}_t, t, \cdot)$. The score model takes as input a sequence of four stacked elements:
\begin{itemize}
\item $\bm{z}_{t}$: the noisy embedding at time $t$, for which we set the rows $\bm{z}_{tn} = 0$ where $\bm{m}_{n}=1$,
\item $\hat{\bm{z}}^{t +1}_{0}$: the estimate of the original noisy embedding at time $t + 1$,
\item $\bm{m}$: a binary mask of length $N$ of which $50\%$ of values are randomly set to $1$, indicating if a location is given ($\bm{m}_n=1$) or is to be generated ($\bm{m}_n=0$),
\item $\texttt{EMB}(\bm{y}_0)$: the original embedding matrix, where we analogously set the rows $\texttt{EMB}(\bm{y}_0)_{n} = \bm{0}$ where $\bm{m}_{n}=0$.
\end{itemize}
Before project the elements through the transformer, we it through an MLP with $[256, 256, 16]$ layers of which the last layer corresponds to the the embedding dimensionality found through an ablation study (Section~\ref{sec:model}). We then apply the conventional sinusoidal embedding from \cite{vaswani2017attention} before projecting the embedded data through a pre-LayerNorm transformer. The transformer uses a self-attention mechanism with four heads. When then conduct the time conditioning using a linear $16$-dimensional layer on the time embedding and adding it to the previous output. The time embeddings are computed using a MLP with two layers of $[256, 256]$ neurons. The time-conditioned embeddings are then fed through an MLP with three layers of $[512, 512, 16]$ nodes. Figure~\ref{fig:architecture} describes the network in greater detail. All three MLPs use \texttt{swish}, or \texttt{SiLU}, activation functions between the hidden layers which we found to be numerically stable independently of the initialization of the neural network weights \citep{hendrycks2016gaussian}. The model has been implemented using the deep neural network library Haiku \citep{haiku2020github}.

\subsection{Model training}

We train the model by creating a data set consisting of sub-sequences of length $N=32$ as input length (which we chose more or less arbitrarily based on the sample size of the data). The model is trained on mini-batches of size $64$ until convergence which we evaluate using a separate validation set that consists of $5\%$ of the entire data. We use an AdamW optimizer \citep{loshchilov2018decoupled} implemented in the gradient processing library Optax \citep{deepmind2020jax} with a linear decay learning rate starting from $lr_1 = 0.0003$ to $lr_S=0.00001$ over $S = 10\ 000$ transitions (i.e., gradient steps). The optimizer uses the default parameterization of $b_1= 0.9$ and $b_2 = 0.99$, and and weight decay of $w_d = 10^{-8}$ (varying of which did not yield significantly different performance results). 

\subsection{Embedding vector weight normalization}

In order to prevent the embeddings to collapse due to jointly learning the embeddings and the score model parameters, we normalize them by their L2 norm before using the embeddings for diffusion. Concretely we implement $\texttt{EMB}(\bm{y}_0)$ by first computing the word embedding matrix of a sequence of locations $\bm{y}_0$ and then normalizing each row, i.e., embedding vector, by its L2 norm.

\subsection{Data}

We use longitudinal GNSS tracking data derived from the GC study conducted by the Swiss Federal Railways (SBB) from November 2016 to December 2017, aiming at evaluating the impact of a mobility offer on individuals' mobility behavior~\citep{martin2019green}. 
The data set comprises $\sim 200$ million GNSS measurements, enabling continuous tracking of $139$ individuals. The median time interval between two consecutive GNSS recordings is 13.9 seconds. 
% stay points
Raw GNSS records undergo preprocessing to identify and delineate stationary areas of individuals, referred to as \textit{stay points}. 
% locations
Subsequently, these stay points are aggregated spatially to create \textit{locations}, denoted as $y_n$, to account for GNSS recording errors that may occur when individuals visit the same place.
% user selection
For further analysis, we restrict our focus to individuals with extended tracking periods ($> 300$ days) and substantial temporal tracking coverage (whereabouts known for $> 60\%$ of the time). Consequently, we retained 93 individuals for the subsequent analysis totalling roughly $45\ 000$ location visits.

We aggregated the data such that consecutive GPS measurements that are within a radius of $100$ meters are denoted as a single location $y_n$. 

Due to privacy reasons, the observational and synthetic data cannot be shared.

\section{Additional experimental results}
\label{appendix:additional-results}

\subsection{Source code}

Source code in the form of a Python package to train a discrete DPM for ILT simulation is available on \href{https://github.com/irmlma/mobility-simulation-cdpm}{GitHub}.

\subsection{Classifier-free guidance}

In addition tot he previously introduced model properties, we experimented with employing \textit{classifier-free guidance} \citep{ho2022classifierfree} to boost the quality of conditionally generated samples. 

During training, classifier-free guidance randomly, i.e, with some propability $p_\text{disc}$ discards the "conditioning" variable as input to the score model. Specifically following the recommendation in \cite{ho2022classifierfree}, with probability $p_\text{disc} = 0.2$, we remove the conditioning arguments of the score model, i.e., we predict $s_\phi(\bm{z}_t, t, \hat{\bm{z}}^t_0, \texttt{EMB}(\bm{y}_0) = \bm{0} , \bm{m} = \bm{0})$ , by that removing both mask and embedding. 

During sampling, classifier-free guidance makes a prediction by taking a linear combination

\begin{equation*}
\hat{\bm{z}}^t_{0}  \leftarrow (1 + w) \cdot s_\phi \left( \bm{z}_t, t, \hat{\bm{z}}^{t+1}_0, \texttt{EMB}(\bm{y}_0), \bm{m} \right) - w \cdot s_\phi \left(\bm{z}_t, t, \hat{\bm{z}}^{t+1}_0, \bm{0}, \bm{0}\right)
\end{equation*}

where $w$ is a hyper-parameter that sets the strengths of the guidance. Ho et al.~\citep{ho2022classifierfree} note that the parameter $w$ essentially trades-off variance of the generated samples and their quality, i.e., Frechet inception distance and inception score which is conventient in practice, since it allows to draw from different conditional distributions after training: setting higher values for $w$ will allow the user to draw samples with higher quality, i.e., closer to the training distibrution and less variance, while lower values for $w$ do the opposite.

In our case, classifier-free guide did however not achieve significant performance improvements.

\subsection{Model ablations}

Below, we present the results of additional ablation studies, i.e., on the choice of the noise schedule and on the choice of the time embedding dimensionality.

\begin{figure}[h!]
\centering
\begin{subfigure}[b]{0.67\textwidth}
\caption{Ablation study on the noice schedule. In comparison to \cite{li2022diffusion} we find that the cosine noise schedule by \cite{nichol21improved} outperforms the sqrt and linear noise schedules.}    
\includegraphics[width=0.8\textwidth]{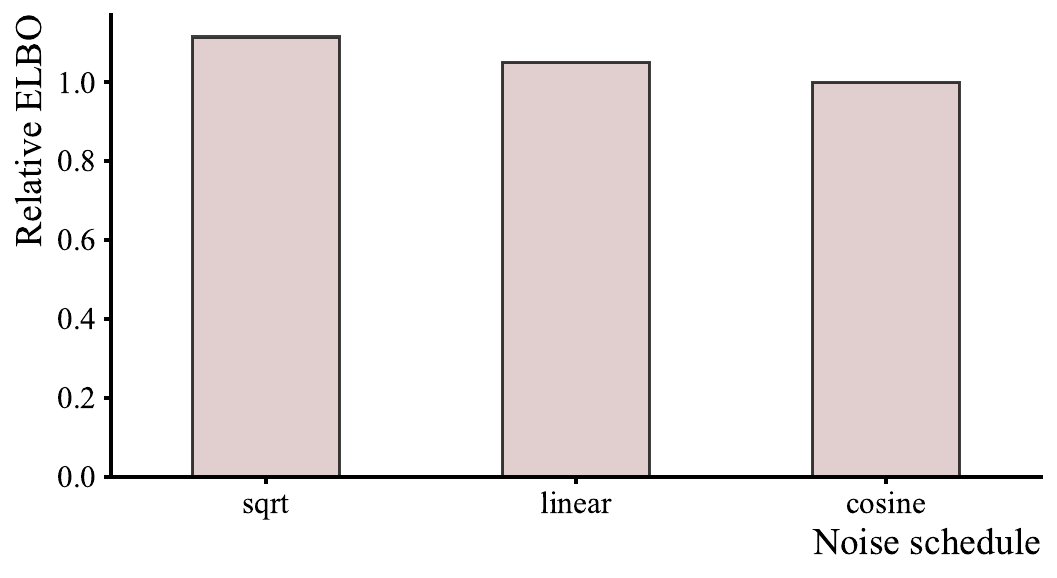}
\end{subfigure}
\begin{subfigure}[b]{0.65\textwidth}
\caption{Ablation study on time embedding. While lower-dimensional time embeddings have roughly the same performance, the dimensionality of $256$ has a slight performance gains for our application (in comparison to the results by \cite{dieleman2022continuous} who used a dimensionality of $128$).}
\includegraphics[width=0.8\textwidth]{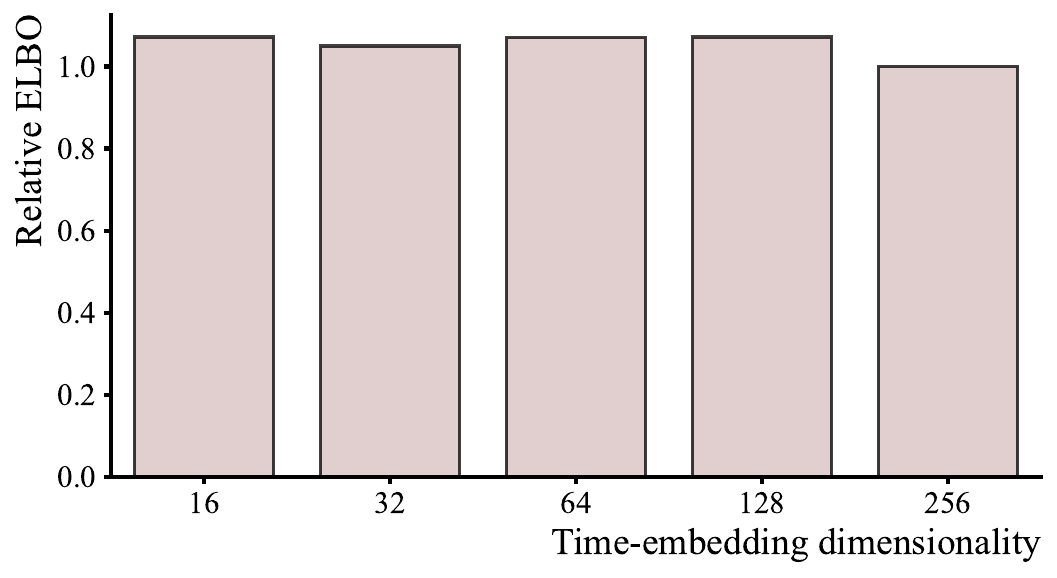}
\end{subfigure}
\end{figure} 

\end{appendices}

\end{document}